# Forbidden knowledge in machine learning
## Reflections on the limits of research and publication

Dr. Thilo Hagendorff
thilo.hagendorff@uni-tuebingen.de
University of Tuebingen
Cluster of Excellence "Machine Learning: New Perspectives for Science"

**Abstract** – Certain research strands can yield "forbidden knowledge". This term refers to knowledge that is considered too sensitive, dangerous or taboo to be produced or shared. Discourses about such publication restrictions are already entrenched in scientific fields like IT security, synthetic biology or nuclear physics research. This paper makes the case for transferring this discourse to machine learning research. Some machine learning applications can very easily be misused and unfold harmful consequences, for instance with regard to generative video or text synthesis, personality analysis, behavior manipulation, software vulnerability detection and the like. Up to now, the machine learning research community embraces the idea of open access. However, this is opposed to precautionary efforts to prevent the malicious use of machine learning applications. Information about or from such applications may, if improperly disclosed, cause harm to people, organizations or whole societies. Hence, the goal of this work is to outline norms that can help to decide whether and when the dissemination of such information should be prevented. It proposes review parameters for the machine learning community to establish an ethical framework on how to deal with forbidden knowledge and dual-use applications.

## 1 Introduction

Currently, machine learning research is, not much different from other scientific fields, embracing the idea of open access. Research findings are publicly available and can be widely shared amongst other researchers and foster idea flow and collaboration. Amongst the description of research findings, scientists frequently share details about their own machine learning models or even the complete source code via platforms like GitHub, GitLab, SourceForge and others. However, the tenet of providing open access contradicts precautionary efforts to prevent the malicious use of machine learning applications. Some applications can very easily be used for harmful purposes, for instance with regard to generative video or text synthesis, personality analysis or software vulnerability detection. Concrete examples for high-stakes machine learning research are OpenAI's GPT-2 text generator (Radford et al. 2019b), Michal Kosinski's and Yilun Wang's software for detecting people's sexual orientation based on facial images (Kosinski and Wang 2018), developments in the field of synthetic media (Chesney and Citron 2018) and many more.

Therefore, researchers have to answer the question whether "AI development [should] be fully open-sourced, or are there ethical reasons to limit its accessibility?" (Murgia 2019) Scientists demand for a "more nuanced discussions of model and code release decisions in the AI community" (Radford et al. 2019a), ask "whether sometimes it is appropriate to withhold open publication of some aspects of [the] research" (Ovadya and Whittlestone 2019, p. 2) and perceive "publication challenges inherent to some parts of AI research" (Clark et al. 2019). Further, machine learning practitioners call for more research on the "efficacy of policies for restricting release of AI technologies" (King et al. 2019, p. 26) and see an "urgent need to develop principled decision-making frameworks" (Solaiman et al. 2019a, p. 15) with regard to publication norms. If machine learning research can possess, as Michal Kosinski admits, "considerable negative implications", which could "pose a threat to an individual's well-being, freedom, or even life" (Kosinski et al. 2013), the decisive question is: Is it worth publishing those research results? Are the reputation gains a researcher achieves by publishing intriguing research insights or the fact that new insights can usher scientific progress worth the risk of others weaponizing those insights? This paper is supposed to be a response to those questions, demands and challenges. Information about or from maliciously used machine learning applications may, if improperly disclosed, cause harm to people or organizations. In this context, harm stands for negative outcomes like fatal or nonfatal casualties, physical and psychological suffering, as well as social, political or economic disruption or damage. The goal of this paper is to outline norms that can help to decide whether and when the dissemination of such information should be prevented.

Without a doubt, research on machine learning has an immense impact on societies. Roughly speaking, the downsides of recent innovations range from threats to democracy and public discourse to massive violations of privacy rights to autonomous weapon systems and algorithmic discrimination (Knight and Hao 2019). This is why perceiving machine learning research solely from the perspective of technological and societal progress is not enough. In fact, machine learning research can have devastating consequences on a grand scale. Of course, the emergence of negative effects of new technologies can rarely be attributed to a single scientific paper. Nevertheless, at regular intervals research findings are published, which can be anticipated with great certainty to be directly abused and to cause damage. Hence, the aim of this paper is to define some initial criteria that can help decide whether research results in the field of artificial intelligence or machine learning should be publicly shared or kept secret or at least under control to a certain extent.

In case of the latter, I want to use the term "forbidden knowledge", which refers to knowledge that is considered too sensitive, dangerous or taboo to be produced or shared (Kempner et al. 2011; Dürrenmatt 1980). Forbidden knowledge is akin to what Nick Bostrom calls an "information hazard", which he defined as "a risk that arises from the dissemination or the potential dissemination of (true) information that may cause harm or enable some agents to cause harm." (Bostrom 2011, p. 2). The term "forbidden knowledge" is used not just to descriptively refer to negative or non-knowledge, but to normative limits of what should be (publicly) known and not known. In this context, it must be considered that forbidden knowledge is a dynamic concept. One cannot determine universally and independently from a certain ethical standpoint, point in time or culture which knowledge should be forbidden. Nevertheless, this paper is intended to assume that, in the light of current political and cultural climates, specific risks are associated with the disclosure of certain research findings in the field of machine learning and that those risks need to be minimized. Hence, it is the social environment that shapes the decision not to make certain knowledge publicly available, not the knowledge itself or certain static categories.

In the first part, this paper outlines the theory of forbidden knowledge in the context of machine learning applications (chapter 3). In the second part, it describes how the emergence, availability and use of forbidden knowledge can be monitored (chapter 4). Various areas in which forbidden knowledge occurs are analyzed. In the last part, methods to control or govern forbidden knowledge are conceptualized, ranging from information sharing rules to pre-publication risk assessments or technical restrictions (chapter 5). The overarching goal of the paper is to propose review parameters for the machine learning community to establish an ethical framework on how to deal with forbidden knowledge.

## 2   Method

Since not much can be known about what is not known publicly, theoretical as well as empirical research on forbidden knowledge is quite difficult. Nevertheless, something can be said about the theory of forbidden knowledge, how to monitor it, in which areas it is produced and how it may be governed. In order to do so, the article is underpinned by a literature analysis. Using search engines like Google Scholar, Web of Science, Scopus, PhilPapers, arXiv and SSRN helped to identify the relevant literature on forbidden knowledge, governance of dual-use technologies – which is especially prevalent in the field of the biology and computer security community – and machine learning risks. Since the literature, especially about forbidden knowledge, is quite broad and heterogeneous, the term "forbidden knowledge" has to be defined, which shall be done in the following chapter. Not only scientific literature was consulted in order to find tangible instances of forbidden knowledge in machine learning. Since many applications are discussed in news articles and not in scientific papers, news articles were gathered via Google News as well.

## 3   Theory of forbidden knowledge

### 3.1   Types of forbidden knowledge

Forbidden knowledge originates in Christianity's idea of Eve eating an apple from the forbidden tree of knowledge (Johnson 1996). Today, the term stands for scientific knowledge that is too dangerous to inquire or to be disseminated unrestrictedly, since knowledge equates power, which can be abused. Hitherto, forbidden knowledge was mostly relevant in disciplines such as IT-security, armaments research and a few other research fields (Tucker 2012a; Freitas 2006). Especially in the field of nuclear physics research, scientific knowledge can become a "mixed blessing", since it "makes the destruction of human life possible" (Smith 1978, p. 34). Moreover, this also applies to other fields like synthetic biology, where for instance a paper described the synthesis of poliovirus (Cello et al. 2002), leading to concerns about bioterrorism. Those cases lead to demands that journal editors may "conclude that the potential harm of publication outweighs the potential societal benefits" (Atlas et al. 2003, p. 1149) and consequently do not publish specific papers at all or only in a modified version (Selgelid 2007).

One can distinguish different types of forbidden knowledge. Knowledge can be forbidden because of the "nature" of the knowledge in question – this is mostly relevant in a religious context –, because of the methods necessary to gain the knowledge – for instance via unethical human experiments – or because of the consequences, which the public availability of the knowledge would have. In the context of this paper, only the latter type of forbidden knowledge shall be considered. In order to narrow down the term even further, only that kind of forbidden knowledge will be discussed for which machine learning techniques are required as an essential factor. Here, only those techniques that are currently available and feasible, omitting to speculate about potential future applications, will be taken into consideration. Furthermore, forbidden knowledge does not only refer to knowledge that is the output of particular machine learning applications – for example classifications about intimate traits of individuals. It also refers to the knowledge about techniques or methods, which are the foundations to build machine learning applications that can be misused – for instance methods to build powerful generators for synthetic media.

## 3.2 Intentions

The concept of forbidden knowledge is closely associated with the theory of dual-use technologies. Concerning the dual-use character of many machine learning applications, one has to distinguish between machine learning scientists who initially undertake the research leading to those applications and users who just utilize them. In this context, the intentions of scientists, which influence the design of the applications, can deviate from the intentions of the users, who can use the applications for purposes other than those for which they were designed. A common scenario in this context would be that researchers pursue good intentions, whereas users have, in many cases, malevolent goals. However, it would be too easy to simply differentiate between good and bad intentions, where the former proceed with dual-use technologies or forbidden knowledge, respectively, in a responsible manner and the latter not. Intentions become bad when they are viewed from the perspective of a certain ethical theory. And even if intentions are identified as bad, this does not mean that the ones possessing them are immoral people.

In this sense, bad intentions are virtually almost the product of certain social situations or systems (Zimbardo 2007). That being said, particular social situations or systems can give rise to intentions to (mis-)use forbidden knowledge to cause harm as well as to motives to prevent forbidden knowledge to come into existence in the first place or to be disseminated in a way that could threaten other individuals or societies. On the one hand, one must assume that there will always be groups of people who seek to enclose forbidden knowledge and prevent scenarios of harm. On the other hand, one must at the same time assume, and this is a fundamental premise of the paper, that there will always be groups of people at private companies, government agencies, research institutes or universities as well as non-institutional entities or private persons who have clearly malign motives and who will try to exploit all that is technically possible in order to pursue their interests. This constant of the existence of malign intentions has to guide the considerations about the responsible handling of forbidden knowledge.

# 4 Monitoring forbidden knowledge

Monitoring the emergence of forbidden knowledge is a prerequisite for the successful governance of it. This means studying emerging machine learning technologies in different areas and sectors and identifying the dual-use characteristics of potentially harmful applications. Following pre-existing decision frameworks for dual-use technologies in other areas (Tucker 2012c; Miller 2018), one can differentiate between the magnitude of potential harms resulting from forbidden knowledge, the imminence of potential harms, the ease of access to forbidden knowledge, the amount of skills needed to gain forbidden knowledge and the awareness about the emergence or malicious use of forbidden knowledge (see Table 1). Besides this categorization of forbidden knowledge in machine learning research, further questions to assess potential publication risks can be asked, i.e. whether harms have the form of structural risks or direct consequences for individuals, other living entities, the environment or non-living things, what type of harm is imminent, i.e. whether it is aiming at physical or mental health, economic stability, human rights, the environment etc., how high the likelihood of the occurring of a particular harm is, whether it is ephemeral or permanent, what possibilities of responding to a specific harm exist, whether the source of harm is traceable and whether potential harms can be redressed (Crootof 2019).

| | low | high |
|---|---|---|
| **magnitude of harm** | single individuals are affected | whole societies are affected |
| | minor detriments, harms or affliction | major detriments or casualties |
| | low monetary costs associated with incidents | high monetary costs associated with incidents |
| **imminence of potential harms** | individuals or organizations can gain and use forbidden knowledge in the remote future | individuals or organizations already possess forbidden knowledge or can gain and use it in the near future |
| **ease of access to forbidden knowledge** | existence of particular kinds of forbidden knowledge is known | knowledge is classified and under control of specialized authorities or closed organizations |
| | technological requirements or knowledge itself are commercially available | hardware, software and data sets are difficult to obtain |
| **amount of skills needed to gain forbidden knowledge** | individuals with low experience or expertise levels can gain forbidden knowledge | only highly skilled individuals can gain forbidden knowledge |
| | ready-made software bundles or how-to manuals exist | advanced persistent threats exist |
| **awareness about the emergence or malicious use of forbidden knowledge** | emergence or malicious use of forbidden knowledge was unforeseen but foreseeable or completely unforeseen | emergence or malicious use of forbidden knowledge is easily discernable |

*Table 1 - Assessment of forbidden knowledge*

## 4.1 Availability

The decisive question is: how difficult is it for individuals with malicious intentions to weaponize a machine learning application in practice? They need to be aware of certain technological features, the necessary skills as well as the necessary resources. The last two aspects depend on the availability of ready-made products or platforms. The more difficult it is to reproduce a certain technical capability in light of the available papers, codes, models, data sets, hardware etc., the more talent, skill or resources are necessary to utilize the capability. But since talent as well as advanced skills are scarce, the likelihood of abuse scenarios drops. Although, in fact, the opposite trend prevails: the likelihood of abuse scenarios increases. While many institutions like OpenAI, Facebook, Microsoft or others do not publish the full-size model of hazardous applications in the first place, freely accessible Internet platforms are emerging elsewhere, providing, in a simplified way, exactly those services that should be kept away from the public.

With "Grover", the Allen Institute for Artificial Intelligence offers a platform where fake news can be created on any given topic (Zellers et al. 2019). The same holds true for a platform from Adam King, which can be accessed via www.talktotransformer.com, or for the language model HAIM of the AI startup AI21 Labs, which can be accessed and used via www.ai21.com/haim. The organization Lyrebird offers the possibility to create a voice recording from any given text (for security reasons up to now only with one's own voice) on its website. Furthermore, and given the case that the software becomes freely purchasable, everyone can edit audio files or fake voices, provided that only a few minutes of voice recordings are available as training data, using Adobe's program "VoCo". On thispersondoesnotexist.com, a software from Nvidia creates deceptively real facial images of people who do not exist. Furthermore, with the application "FakeApp" anyone can create DeepFakes.

Hence, it seems obvious that technologies that can be instrumentalized for modern disinformation campaigns are becoming more widely available. It is to be expected that a process of "deskilling", meaning a gradual decline in the amount of needed knowledge or the rise of ready-made software bundles or how-to manuals, affects the use of machine learning methods and applications, with the result that the number of individuals who can use potentially harmful machine learning technologies grows constantly. This process of "deskilling" is combined with a "democratization" of machine learning and its requirements like training data sets or software libraries. Open access and open source give rise to many innovations while rendering tools freely available to bigger and bigger crowds. At the same time the number of (non-expert, amateur) individuals with potentially malign intentions and willing to reinterpret the purpose of dual-use technologies in a harmful context grows.

Possibly, the mere idea of using or processing data with methods of machine learning in a certain way poses the threat that individuals with malicious intentions realize this particular idea themselves. This is what Nick Bostrom (2011) calls "idea hazard". The "idea hazard" is accompanied by the "data" or "product hazard" (Ovadya and Whittlestone 2019), where machine learning applications themselves or their respective outputs pose a danger. All the mentioned hazards can be conveyed by openly accessible research papers, containing more or less details about particular software solutions. Open access, however, is not a "binary variable" (Brundage et al. 2018, p. 87). There are gradations in the way research results can be shared, for instance by adopting a staged publication release during which various (negative) effects of the released applications are monitored. Tangible information sharing rules relate to the kind of information or code that is exchanged or selectively made public. Documents about applications possessing certain risks of causing harm when used "in the wild" in an uncontrolled manner can contain rough descriptions of the achievement or simple proofs of concept. A next step is to publish pseudocode, parts of the code or machine learning algorithms without the necessary hyperparameters. Ultimately, papers can contain appendices with fully working exploits or the complete code together with the trained model as well as tutorials (see Table 2).

Each level of sharing information raises the risk that third parties can use the research for malicious or criminal purposes. The fewer details that are shared by researchers, the higher the need for technical expertise on the part of third parties who wish to reproduce or harness the original achievements. Apart from the extend of shared technical details, the availability or reproducibility of forbidden knowledge depends on many further conditions: the amount of monetary costs to acquire certain hard- or software, whether code is equipped with or without comments as well as their level of detail, whether code is compiled or raw, whether details about the types of the required hardware being used are known and so forth.

| availability | low | | | | | | | high |
|---|---|---|---|---|---|---|---|---|
| **levels** | mere idea or concept | concept paper | pseudo code | detailed paper | training data | trained models | source code | ready-made product |
| **further dimensions** | amount of monetary costs to acquire particular sets of hard- or software / code equipped with or without comments / code equipped with or without hyperparameters / code is compiled or raw / details about the types of hardware being used | | | | | | | |

*Table 2 - Grades of availability*

## 4.2 Areas of forbidden knowledge

In the following section, the paper sheds light on different areas of forbidden knowledge. Examples of machine learning based applications that were retracted or never got published are to be described. Various fields of application will be discussed, including contexts like sexuality, social manipulation, algorithmic discrimination, artificial general intelligence as well as further areas where sensitive information can be produced or acquired.

Regardless of the particular area, one can differentiate between machine learning applications that aim at single individuals (e.g. "gaydar" applications) and applications that aim at a wider social context or whole societies (e.g. artificial general intelligence). Furthermore, one has to distinguish between single inventions or individual research works that result in forbidden knowledge (e.g. research on automatically winning commercial games) and the dynamics of many small inventions or consecutive research activities that gradually produce forbidden knowledge (e.g. research on deep fakes). Moreover, applications that gather or detect sensitive information (e.g. digital suicide risk detectors) have to be differentiated from applications that generate or make up fake sensitive or discrediting information (e.g. text generators). In addition to that, either information on machine learning applications itself or information in the form of an application's output can be considered to have the status of forbidden knowledge.

### 4.2.1 Synthetic media

Research on synthetic media and the publication of corresponding findings and insights is delicate. This was perfectly shown by researchers at OpenAI. They developed a text generator called GPT-2 (Radford et al. 2019b) which is so powerful that they decided to follow a staged release policy, meaning that they released a small model with 124 million parameters in February 2019, a model with 355 million parameters in May 2019, a model with 774 million parameters in August 2019 and the full model in November 2019 (Radford et al. 2019a; Clark et al. 2019; Solaiman et al. 2019b). OpenAI has teamed up with several partner universities studying the human susceptibility to artificially generated texts, potential misuse scenarios or biases in the produced texts. The original decision not to release the full-fledged text generator was fueled by fears concerning the circumstance that GPT-2 could significantly lower the costs of disinformation campaigns or simplify the creation of spambots for forums or social media platforms. Although admitting that they found "minimal evidence of misuse" via their "threat monitoring" (Solaiman et al. 2019a), doubts are justified as to whether their monitoring, which was mainly focused on online communities, was really reliable, since OpenAI and their partners can mostly monitor current public plans or cases of misuse for their application, but not non-public or potential future misuse scenarios as well as advanced persistent threats.

In the overall view, machine learning technologies make it possible to automatically create any media, be it images (Karras et al. 2017), videos (Thies et al. 2016), audio recordings (Bendel 2017) or texts (Radford et al. 2019a). The quality of the media created is constantly improving, so that previously accepted principles such as "seeing is believing" or "hearing is believing" have to be abandoned. Whether the content corresponds with actual events does not matter. While researchers try to catch up and find solutions to reliably detect fake samples produced by generative adversarial networks (Valle et al. 2018; Rössler et al. 2019), it remains true that generative models make it really easy to generate or edit media. Despite those technical solutions to detect synthetic media and approaches to educate humans on detecting machine manipulated media (Groh et al. 2019), a further, quite strict idea is to limit the availability of trained generative models. Against this background it is astounding how unquestioningly papers have been published in recent years, in which leap innovations in the generation of fake media, especially videos, are described – although many research groups, for instance the one behind Face2Face, did not release their code (Thies et al. 2015; Thies et al. 2016; Thies et al. 2018; Fried et al. 2019; Ovadya and Whittlestone 2019; Thies et al. 2019). Synthetic videos, no matter if they are generated through Face2Face, DeepFakes, FaceSwap or NeuralTextures, can have all sorts of negative consequences, from harm to individuals, national security, to the economy and democracy (Chesney and Citron 2018). Fake porn is used to intimidate journalists, fake audios to mimic CEOs and commit fraud, fake pictures to trick other people into disclosing sensitive information (Harwell 2018; Satter 2019) and so on. Despite obvious risks, the improvement of synthetic media also poses the risk of people claiming that real footage is fake, erroneously denying its verisimilitude. In this context, technical solutions to detect synthetic media should also operate in the opposite direction, meaning that they should be able to detect recordings that are real.

### 4.2.2 Social manipulation

Issues of social manipulation via machine learning applications came especially into public awareness after reports about the role of Cambridge Analytica during the successful UK's Vote Leave campaign and the 2016 US presidential election. Some of the methods used during the elections trace back to research in the field of psychometrics. Here lies the origin of methods where individual's psychological profiles are automatically extracted from their (harmless) digital footprints via machine learning in order to influence their behavior or attitudes. Researchers proved that very few data points a particular individual generates suffice to make accurate predictions about personality traits (Youyou et al. 2015; Kosinski et al. 2015; Lambiotte and Kosinski 2014; Kosinski et al. 2014), which can in turn be used for improved persuasion techniques, called "micro targeting". Micro targeting, for instance, can significantly raise the click-through-rates of personalized online advertisements (Matz et al. 2017). However, it is not just advertisements. Several companies exploit techniques where psychometrics and machine learning are combined in order to conduce "behavioral change programs", blurring the lines between the military and civic use of (dis-)information campaigns or "psy-ops" (Ramsay 2018). The scientists involved in the related research honestly talk about "considerable negative implications" (Kosinski et al. 2013) of their work. Psychometrics research, which builds the foundation for methods of social manipulation, was metaphorically called a "bomb" (Grassegger and Krogerus 2016). Others dubbed the methods used by Cambridge Analytica and similar organizations "weapons-grade communication techniques" (Cadwalladr 2019), which clearly points at the dangers certain machine learning applications can pose. If the scientists, whose research builds the foundation for micro targeting and other methods of machine learning based social manipulation, had not made their work publicly accessible or not carried it out in the first place, they would have been deprived of a great deal of reputation. At the same time, however, it can be assumed that numerous negative effects could not have taken place, for example with regard to election rigging.

### 4.2.3 Discrimination

The emergence of (unintended) algorithmic discrimination can be a reason for retracting machine learning applications. To name just two examples: in 2016, Microsoft developed a chat bot called "Tay". It was released on Twitter, so that users could have a conversation with it. After a while, it was bombarded with racist, sexist language by trolls. And since "Tay" is based on machine learning algorithms, it inherited and automatically reproduced the discriminatory language (Misty 2016). After one day, Microsoft had to retract the application. This example shows the danger of not anticipating that machine learning applications can be manipulated by adversarial inputs or not anticipating to equip those applications with meta rules, meaning that programmers define boundaries, software agents are not allowed to overstep (Wallach and Allen 2009). Another example where failures to prevent discrimination were an issue and lead to the retraction of machine learning based tools is Amazon's experimental hiring software. The software used machine learning techniques to score job candidates (Dastin 2018). It discriminated against women, since it was trained with patterns in applications that were submitted over a 10-year period, which came mostly from men. Amazon had to shut down the project after they found out about its shortcomings.

Those two examples, to which many more could be added (Bolukbasi et al. 2016), depict cases of discrimination through technology. Another strand of discrimination occurs when technology is used to assist discrimination. One can for instance think of using machine learning methods to sift through data sets containing demographics, profiling, biometric, medical or other behavioral data in order to gain insights about racial, sexual or cognitive differences between different groups of persons. It must be stressed that in this context, research on the nature versus nurture debate can be very problematic not only due to potentially malicious interest of the involved researchers, but also due to the inability of the public to deal with corresponding research findings and due to the political consequences the publication of particular findings would likely have. The same holds true with regard to machine learning based research on mental illnesses or intelligence. For instance, researchers showed that social media profiles or especially Facebook posts can be used to predict depression (Choudhury et al. 2013; Eichstaedt et al. 2018). Those insights can be used for the common good, but also for purposes of unjust social sorting or discrimination (Lyon 2003).

### 4.2.4 Sexuality

Issues related to sexuality are another area where machine learning applications can cause widespread harm. For instance, a software called "DeepNude" found rapid sales, allowing users to automatically render pictures (of women) into nude photos. Shortly after its release, the developers stopped offering the software (Quach 2019), but one can still use it via various online platforms. Deciding to stop selling the software did obviously not stop its further dissemination. Another case, where a particular machine learning application is more than just prone to abuse, is the use of deep neural networks for detecting sexual orientation from facial images. This was first demonstrated in the famous paper by Michal Kosinski and Yilun Wang (2018). The study raised a lot of criticism (Todorov 2018), but its findings were later confirmed by a replication study (Leuner 2019). John Leuner, who conducted the replication study, claimed that the research "may have serious implications for the privacy and safety of gay men and women" (Leuner 2019, iii), a sentence which is nearly identical with the claim of Kosinski and Wang, who write that their findings "expose a threat to the privacy and safety of gay men and women." (Kosinski and Wang 2018) To prevent this threat to a certain degree, Leuner did not disclose the source of his data, which he collected for his study. Otherwise, Kosinski and Wang stress that abandoning the publication of their findings "could deprive individuals of the chance to take preventive measures and policymakers the ability to introduce legislation to protect people." (Kosinski and Wang 2018) The researchers hope that upcoming or current post-privacy societies are "inhabited by well-educated, tolerant people who are dedicated to equal rights" (Kosinski and Wang 2018). This may sound naïve, especially in view of current political trends and raising group-focused enmity. Therefore, the misuse of the aforementioned research is a considerable concern, advising stronger caution when publishing research results in the context of machine learning applications dedicated to reveal or generate traits connected to sexuality or sexual orientation.

### 4.2.5  Further sensitive fields

What holds true for sexuality is at the same time applicable to further sensitive fields where machine learning techniques are applied to detect forbidden knowledge about an individual's intelligence, political views, ethnic origin, wealth, propensity to criminality, religiosity, drug use or mental illnesses. Regarding the latter, Facebook, for instance, repeatedly ushered initiatives for suicide and self-harm prevention. By merely analyzing likes, comments or other interactions on their platform, Facebook can "sense" suicide plans and help affected persons or persons who are related to the affected ones via overlays with information on suicide prevention. Due to its sensitive nature, this tool was not released in Europe and is therefore representing another case of forbidden knowledge (Keller 2018). Information for instance about one's suicide risk are traditionally protected by privacy norms (Veghes et al. 2012, p. 705). Those norms were first and foremost based on restricting access to or controlling the dissemination of personal information, for example via concepts of contextual integrity (Nissenbaum 2010; Tavani 2008). With regard to existing machine learning techniques, those methods are obsolete (Belliger and Krieger 2018; Hagendorff 2019a). Now, intimate personal information like sentiments or personality traits can not only be automatically extracted from Social Media profiles (Youyou et al. 2015), but also from personal websites or blogs (Marcus et al. 2006; Yarkoni 2010), pictures (Segalin et al. 2017), smartphone usage (Cao et al. 2017; LiKamWa et al. 2013) and many more. Furthermore, particularly sensitive applications for purposes of reading one's mind, for rudimentary brain-to-brain interfaces or even the decoding of dreams are being developed (Jiang et al. 2019; Horikawa and Kamitani 2017). This new, machine learning based research stands in a long tradition of trying to control, read or manipulate individual's minds with different technologies (Wheelis 2012).

Apart from machine learning technologies which aim at single individuals, there are applications that have effects on a more general, societal level. Examples for such applications, that can likewise fall under the category of forbidden knowledge, could be innovative artificial trading agents used for market manipulation (Wellman and Rajan 2017), software to conduct automated spear phishing (Seymour and Tully 2016), "AI 0-days" or other massive vulnerabilities in machine learning procedures itself, as well as methods for automated software vulnerability detection (Brundage et al. 2018), classified surveillance technologies, the combination of data from fleets of earth-observing satellites with news sources, mobile devices, social media platforms and environment sensors (Kova 2019) or even autonomous applications build to assist with or conduct torture (McAllister 2017). In addition to such rather obvious areas where forbidden knowledge may occur, machine learning applications have also been held back in less obvious places as a result of risk assessments. For instance, a 2019 publication (Brown and Sandholm 2019) demonstrated how "Pluribus", an AI based software, is stronger than professional human players in six-player no-limit Texas hold'em poker.

With a short reference to the fact that the "risk associated with releasing the code outweighs the benefits" (Brown and Sandholm 2019, p. 7), the researchers decided to only release the pseudocode, but not the complete program in order to not harm the poker community, as well as online poker companies. Ultimately, not only in poker, but in any online game where players can win money, it is to be expected that AI applications can be used to win money illegitimately using autonomous agents. The fact that software developers decide not to publish programs in this context is just another symptom of an increasing amount of forbidden knowledge in machine learning.

Moreover, the creation of artificial general intelligence is associated with the fear of technology developing an uncontrollable momentum of its own (Bostrom 2014; Omohundro 2014, 2008; Tegmark 2017). That is why some researchers demand to halt every research effort aiming for artificial general intelligence – despite the fact that discussions around artificial general intelligence are often quite far-fetched and speculative. Nevertheless, this does not mean that current technologies are completely rid of the risk of becoming uncontrollable. When Facebook, for instance, developed an artificially intelligent bot for negotiation purposes (Lewis et al. 2017), it happened that the negotiation software transitioned from using English language to a language or dialect of its own – which humans cannot understand. This phenomenon of autonomous agents developing code words for themselves (Mordatch and Abbeel 2017; Das et al. 2017; Lazaridou et al. 2016) prompted the Facebook researchers to shut down the negotiation bot (Wilson 2017) in order to stay in control about what the system is communicating.

## 5   Governing forbidden knowledge

In stable societies, most tools or technologies that are inclined or designed to harm other people are subject to a certain degree of control. This is most obvious in gun control, but also in areas like biological or chemical weapon conventions. In this context, material goods are affected. However, through machine learning and the related possibilities of harming other individuals or organizations, societies are increasingly entering a situation where similar regulatory measures for certain software or kinds of information need to be considered. This can be explained in particular by the fact that digital goods are easily scalable, which is not the case for material goods. This scalability, together with the observation that the global networking of technical artefacts generates a general loss of control over the distribution of digital goods, makes the new situation an immense technical as well as societal challenge. Information can be multiplied with virtually zero marginal costs and it can be transmitted without any or very few traces. Hence, export controls, reporting requirements as well as other mandatory regulations are very difficult to enforce. In fact, research on information control consistently states that "ontological friction", meaning the forces that oppose the flow and dissemination of information, is constantly decreasing (Floridi 2005, 2006; Hagendorff 2017, 2018, 2019b). Against the backdrop of digitization, the efforts required to generate, obtain, process and transmit information are becoming less and less, with the result that societies are "informationally porous" (Floridi 2010, p. 7).

Notwithstanding these difficulties, however, a number of measures can be considered to control forbidden knowledge (see Table 5). Those measures or methods can comprise soft or hard laws. On the one hand, soft laws are generally accepted, voluntary norms about restrictive information sharing principles without rigorous enforcement. On the other hand, hard laws are mandatory, legally enforceable measures. Following Jonathan Tucker (2012b), one can distinguish between more or less stringent governance measures, ranging from statutory regulations or reporting requirements to security guidelines or pre-publication reviews up to codes of conduct, transparency measures or risk education and general awareness raising with regard to the dual-use nature of machine learning techniques (Minehata and Sture 2010). Although a bunch of international and national governance approaches (Daly et al. 2019) as well as legal norms already exist, regulating complexes like privacy, data protection, security, confidentiality, environmental protection, armament, labeling and many more, numerous areas of machine learning research and development are unregulated and in need of legal enactment (Calo 2017). At the same time, one also has to keep in mind that the enactment of laws which are supposed to deal with complex dynamics in technology development, where potential harms can only be foreseen through vague technology assessment (Collingridge 1980), bears the risk of stifling innovations or smothering promising technologies in an early stage. In the end, governance measures must be able to accommodate to new developments and findings. Norms, laws, principles or rules should be modified or amended in an iterative manner in order to adapt and keep pace to an ever changing and highly accelerated field of research and technological change which constantly creates new risks and benefits.

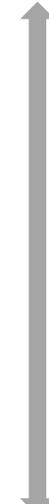

| | measures | potential side effects | |
|---|---|---|---|
| **hard law** | legal regulations | stifling innovations / restricting scientific freedom | more stringent |
| | mandatory registration / reporting requirements | some research may drift off into secrecy | |
| | export controls | emergence of black markets | |
| **soft law** | security guidelines | guidelines are not implemented in practice | |
| | organizational self-governance | not all relevant organizations will participate | |
| | international standards | standards are too abstract and not applicable to specific situations | |
| | pre-publication risk assessment | scientists downplaying risks in order to promote and publish their research | |
| **informal** | codes of conduct | treaties having a "figleaf function", fending off criticism | |
| | information sharing | abuse of confidence of partner organizations | |
| | risk education and awareness raising | empowering malicious agents or organizations | |
| | whistle-blowing channels | whistleblowers must endure repressions of all kinds | less stringent |
| | transparency measures | giving inspirations for abuse scenarios | |
| **technical** | central access licensing models | concentration of power on a few particular organizations | |

*Table 3 - Governance of forbidden knowledge, based on the spectrum of governance measures by (Tucker 2012b)*

## 5.1 Limits on the pursuit of knowledge

A limitation of the approach described in this paper is that restrictions on publication practices in the field of machine learning would not be necessary at all if research directions or strands which can obviously lead to harmful outcomes would not be pursued in the first place – an approach which Seumas Miller (2018) called "collective scientific ignorance". This would also mean not to encourage pursuing those research directions or strands via call for abstracts or papers in the context of conferences or journals. However, machine learning is in most cases a general purpose or dual-use technology, meaning that it has general capabilities, which are applicable to countless varying purposes. Thus, any efforts to restrict foundational research in the field are misleading. The only reasonable way of limiting the pursuit of knowledge at all is to restrict very specific strands of applied research, for instance, in the field of synthetic media – which is again a problematic step since it undermines the scientific freedom.

Despite that, a popular argument which is used to counter the idea to restrict research strands or scientific publication practices is to say that "if we do not do research on X or develop X, someone else will". This argument is used to wipe away all sorts of ethical concerns. However, it fails for several reasons. First, many studies show that moral behavior is "contagious", meaning that the morally desirable as well as the unacceptable behavior swiftly finds imitators and spreads (Kraft-Todd et al. 2018; Bollinger and Gillingham 2012; Eker et al. 2019). Applied to the field of ethics of science, this means that abstaining from pursuing certain research questions can be seen as an example and is imitated, while an "if we do not do research on X or develop X, others will do it" attitude leads to a general lack of accountability and disregard for ethical concerns. Second, the argument fails because of the fact that leap innovations are of course not inevitably made. The correct formula would be: "If we do not do X, then others might do it with a certain likelihood."

## 5.2 Pre-publication risk assessment

Demands for changing the common peer review process in computer sciences with regard to a greater caution for "side effects" of machine learning technologies are nothing new (Hecht et al. 2018). The idea is that computer scientists shall at least be obliged to add paragraphs about all reasonable broader impacts, both positive and negative, to their papers and proposals. In case a research proposal or paper has predominantly negative impacts, it should discuss complementary technologies or other interventions that could mitigate these impacts – or reviewers should be encouraged to simply reject the proposal or paper.

Journal editors could be part of this process of a pre-publication risk assessment (see Table 6). Although polls show that the majority of researchers disagree that journal editors should reject a paper, if the reviewers have concerns about the social acceptability of the research findings (Kempner et al. 2011), this does not mean that the idea of journal editors rejecting papers for security concerns is completely declinable. Having said this, another option editors have is not to reject certain papers, but to demand a staged or delayed publication, so that unintended consequences or misuse scenarios can be scrutinized for the time being. Nevertheless, one must consider two limitations of this approach. First, research also takes place in companies, where no typical academic publication process is entrenched. Second, current publication practices embrace the possibility of releasing papers via preprint servers like arXiv, where no thorough peer-review takes place. For instance, when mathematicians were asked if they would publish a dangerous algorithm, say a solution to break encryption solutions via an algorithm for fast factorization, the typical response is: "'I would publish it on arXiv immediately. It's my right to publish whatever mathematical work I do.'" (Chiodo and Clifton 2019, p. 14) In this view, preprint servers can be an impediment to responsible disclosure practices in the sciences. Responsible disclosure means – to stick to the example of IT security – to delay the publication of vulnerability descriptions until patches or other precautionary measures are in place. But since there might be no "patches" when it comes to the dissemination of certain types of forbidden knowledge in machine learning, it is an ethical requirement that researchers refrain from conducting research or publishing it in certain areas.

Every scientist has interests (Johnson 1999). Especially in the field of machine learning, one can become well known or even famous within one's own community relatively easily, given that there are still numerous unexplored areas within research and due to newly available methods – in particular deep neural networks (Fan et al. 2019) –, breakthroughs and innovations can be achieved in various fields. If scientists or research organizations decide to withhold certain research results from the public or to demonstrate it only to certain colleagues or media representatives, this can mean they are using this move as a marketing strategy. Apart from that, one has to think about how to compensate absent reputation gains in case of self-chosen or externally enforced publication restrictions. Delayed or absent publication means that opportunities for reputations gains or subsequent research are lost in order to mitigate or prevent potential risks. At the same time, however, the scientific reward system or career advancement, which requires a long publication list and pushes scientists to publish their research, can counteract security concerns (Selgelid 2007). Furthermore, not releasing research papers at all or not to the full extent has the disadvantage that it is harder to replicate and, in this way, approve or refute particular research results. A possible solution to mitigate this situation is to establish robust communication channels between machine learning research organizations (Solaiman et al. 2019a), following established information sharing principles, for instance, in the computer security community.

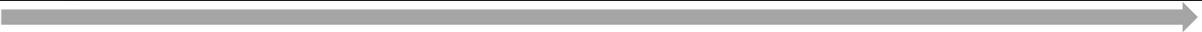

| publication process | | |
|---|---|---|
| time | | |
| immediate release via preprint | pre-publication risk assessment | no publication |
| | | staged publication |
| | | selected distribution |
| | | full publication |
| | normal publication via peer-review | |

*Table 4 - Paths of publication*

## 5.3   Information sharing rules

In the context of information sharing rules, pre-existing considerations from the field of IT security and its practice of exchanging information on cyber threats can be drawn upon (Johnson et al. 2016; Sillaber et al. 2016). The IT security community already has standardizations for cyber security data sharing like, for instance, "Structured Threat Information eXpression" (STIX) or "Trusted Automated eXchange of Indicator Information" (TAXII) as well as more than twenty threat sharing organizations or platforms like the "Malware Information Sharing Platform" (MISP), the "Information Sharing and Analysis Center" (ISAC) or the "Information Sharing and Analysis Organization" (ISAO) at their disposal, where companies can share risk information in a secure environment (Sauerwein et al. 2017). Such an infrastructure is missing for machine learning applications and their potentially harmful societal implications.

Nevertheless, some preliminary ideas for sensitive information sharing rules in the field of machine learning can be drafted. The rules are supposed to set restrictions for sharing and distributing research findings, training data, code or final-trained neural network models. Sharing is only supposed to happen when the trustworthiness of the recipients can be guaranteed. Furthermore, the goals and objectives, as well as the scope of information sharing should be clear. Before sharing takes place, it should be considered whether parts of the information have to be anonymized or sanitized. Ultimately, the potential impacts of information sharing with particular third parties should be assessed. This can either mean not to share information outside of one's own meeting or just verbally and in person. It can also mean to exchange information only inside of one's own organization and with trustworthy clients. Or it can mean to share sensitive information along a predetermined set of trusted partner organizations, which would be the laxest form of a selective sharing regime (Johnson et al. 2016, p. 15).

Moreover, a further idea how to deal with potential harmful machine learning applications in order to restrict access to them, while at the same time making them available to authorized persons, is to provide a central access licensing model (Brundage et al. 2018, 89 ff.). This means that authorized persons can access particular capabilities of a given machine learning application remotely via application programming interfaces (APIs), while the application itself, and respectively its code, is stored at a secure location or storage, similar to a cloud provider. Advantages of such a model lie in the fact that restrictions like limitations to the frequency of the usage of particular capabilities or speed limitations might prevent at least some malicious or abusive use cases. Furthermore, the terms of service can legally prohibit particular use cases. Central access licensing models cannot thwart the abuse of its services in general, but they can at least lower the likelihood of misusage.

## 5.4   Making forbidden knowledge public

Withholding forbidden knowledge or, to be more precise, the existence of certain types of forbidden knowledge from the public is not per se appropriate (Bostrom 2017). Apart from research organizations, this holds true especially for intelligence agencies, which may be ahead of time with regard to the use of certain machine learning technologies and possess ways of acquiring forbidden knowledge without public consent or public awareness (Kova 2019). Since a situation like this can contradict democratic norms, it may be ethically required to speak openly about forbidden knowledge in intelligence agencies in order to raise public awareness and to let democratic representatives, who oversee those agencies, decide whether the used methods are legitimate. Scientists, especially those who are working for intelligence or military agencies, may feel an obligation to speak truth to power and to warn the public about certain threats emerging from machine learning applications. In general, only through public awareness wider societal mitigation measures can take place, for instance new anti-discrimination norms (Hagendorff 2019a), resilience against attacks via synthetic media (Chesney and Citron 2018) or research on technical counter measures (Li et al. 2018). This way of arguing is already established in large parts of synthetic biology, where scientists stress that when making forbidden knowledge public, it is important to inform a wider scientific community about new threats in order to support understanding of those threats and to serve biodefence preparations (Selgelid 2007; Atlas et al. 2003).

## 5.5   Postdisclosure measures

Postdisclosure measures are important, because even if a particular research institute decides not to publish high-stakes machine learning research results, other institutes may make simultaneous discoveries or inventions and choose not to refrain from publication. Furthermore, defectors who do not adhere to non-disclosure agreements of particular organizations may decide to pursue or publish research results which depict forbidden knowledge and gain a competitive edge (Ovadya and Whittlestone 2019). Thus, in many cases it is only a matter of time until the availability of methods to acquire forbidden knowledge is on the rise. Combined with the prospect of less and less possibilities of information control, meaning that mechanisms to govern forbidden knowledge are in many cases highly unreliably, one has to decide: is it better to spend one's resources on restricting the dissemination of forbidden knowledge or should one put effort into preparing the society to deal with high-stakes machine learning applications – or would the best option be to restrain from following particular research questions in the first place? While the former and the latter option cannot be generalized, since it can always be assumed that individuals or organizations with harmful motives exist, efforts to prepare societies for risky applications of machine learning can be generalized freely. This does not mean that only one of these approaches should be pursued. Rather, the combination of abstaining from specific research, restricting the dissemination of knowledge and educating people about risks is promising.

## 6  Conclusion

Discourses about publication restrictions are already entrenched in the field of IT security as well as in biotechnology research. This paper makes the case for transferring this discourse to the field of machine learning research. No different from the aforementioned strands of research, inventions and scientific breakthroughs in machine learning can yield dual-use applications that pose massive threats for individual or public security. This paper is a plea to take those risks more seriously. All in all, a bunch of different methods have to be combined in order to deal with forbidden knowledge in an appropriate manner. Those methods comprise monitoring measures of forbidden knowledge, expert risk evaluation, education in responsible research processes, pre-publication risk assessments, responsible information sharing as well as disclosure rules, technical restrictions, postdisclosure measures and many more. In sum, the idea of forbidden knowledge in machine learning should not put limits or constraints on science or the pursuing of legitimate research questions – but limits on the way research insights are shared. These limits should be established not because machine learning science itself is dangerous. Rather, it is the current political and cultural climate in many parts of the world that brings forth risks of misusing software as a tool to harm or suppress other people.

Having this in mind, a tangible political response not only to the increasing importance of machine learning technologies in more and more areas of life and work, but also to the emergence of forbidden knowledge or highly sensitive information, would have to establish specific authorities to deal with the responsible research, development and application of artificial intelligence. Just as environmental, civil protection or nuclear energy authorities are used to safeguard a country's population, a centralized state artificial intelligence agency could, among other things, institutionalize technology monitoring as well as address the processes of managing forbidden knowledge as described in the text. An artificial intelligence agency could help to keep certain sensitive information classified while, at the same time, educating the public about the existence of forbidden knowledge and potential consequences of its spreading. This education should aim at the promotion of tolerance and equal rights as well as collective moral responsibility on the part of machine learning scientists. As long as such an agency is not put into practice, the task of monitoring and governing forbidden knowledge in machine learning must be shared between teams of researchers, the administration of research institutions, journal editors or other independent committees, where everyone possesses a partial, individual share of the overall collective moral responsibility.

# 7 Acknowledgements

This research was supported by the Cluster of Excellence "Machine Learning – New Perspectives for Science" funded by the Deutsche Forschungsgemeinschaft (DFG, German Research Foundation) under Germany's Excellence Strategy – Reference Number EXC 2064/1 – Project ID 390727645.

## Publication bibliography